\documentclass[fleqn,10pt]{wlscirep}
\usepackage[utf8]{inputenc}
\usepackage[T1]{fontenc}
\usepackage{xcolor}%
\usepackage{comment}
\usepackage{tcolorbox}
\usepackage{verbatim}
\usepackage{float}
\usepackage{enumitem}
\usepackage{subcaption} 

\usepackage{hyperref}
\usepackage{xr-hyper}

\newcommand{\Mytitle}{A Large-Language-Model Framework for Automated Humanitarian Situation Reporting}

\title{\Mytitle}

\author[1]{Ivan Decostanzi}
\author[1]{Yelena Mejova}
\author[1,2,*]{Kyriaki Kalimeri}
\affil[1]{ISI Foundation, Turin, Italy}
\affil[2]{UNICEF, NYC, USA}

\affil[*]{kyriaki.kalimeri@isi.it}


\begin{abstract}
Timely and accurate situational reports are essential for humanitarian decision-making, yet current workflows remain largely manual, resource intensive, and inconsistent. We present a fully automated framework that uses large language models (LLMs) to transform heterogeneous humanitarian documents into structured and evidence-grounded reports. The system integrates semantic text clustering, automatic question generation, retrieval augmented answer extraction with citations, multi-level summarization, and executive summary generation, supported by internal evaluation metrics that emulate expert reasoning.
We evaluated the framework across 13 humanitarian events, including natural disasters and conflicts, using more than 1,100 documents from verified sources such as ReliefWeb. The generated questions achieved 84.7 percent relevance, 84.0 percent importance, and 76.4 percent urgency. The extracted answers reached 86.3 percent relevance, with citation precision and recall both exceeding 76 percent. Agreement between human and LLM based evaluations surpassed an F1 score of 0.80. Comparative analysis shows that the proposed framework produces reports that are more structured, interpretable, and actionable than existing baselines.
By combining LLM reasoning with transparent citation linking and multi-level evaluation, this study demonstrates that generative AI can autonomously produce accurate, verifiable, and operationally useful humanitarian situation reports. 
\end{abstract}

\begin{document}

\flushbottom
\maketitle
\thispagestyle{empty}

\section*{Introduction}
The humanitarian sector operates in complex, high-pressure environments characterized by rapidly changing circumstances, data scarcity, and fragmented information flows. In crisis situations such as natural disasters, conflicts, or health emergencies, organizations must collect, verify, and synthesize information from multiple heterogeneous sources, including government bulletins, NGO publications, news outlets, and internal reports, often under severe time constraints. One of the key outputs of this process is the \textit{situational report}, a critical document that provides a concise snapshot of an ongoing humanitarian crisis, outlining its scope, impact, immediate needs, and ongoing relief efforts~\cite{situation_reports_2024}. These reports are vital tools for coordination and decision-making among governments, NGOs, and international organizations. Yet, despite their importance, there is limited understanding of how artificial intelligence can replicate the analytical reasoning required to synthesize fragmented crisis information.

Producing high-quality situation reports remains a significant challenge. The process is typically manual, time-consuming, and resource-intensive, requiring substantial human effort to filter, summarize, and contextualize information derived from diverse and unstructured data sources~\cite{national2019decadal-SitRep}. Even with such investment, recent findings indicate that many reports, particularly within the UN system, are seldom read by their intended audience~\cite{nobody_reads_un_reports}. This inefficiency underscores a fundamental research question: \textit{can generative language models not only accelerate information synthesis but also preserve the factual reliability and analytical depth required for humanitarian operations?}

Recent advancements in Artificial Intelligence (AI), and particularly in Generative AI, show great promise in addressing these challenges~\cite{alnap_ai_humanitarian_2025}. AI systems can process and analyze large volumes of textual and numerical data, enabling faster and more accurate extraction of key insights from multiple sources. The emergence of large language models (LLMs) has transformed natural language processing (NLP), substantially improving performance in text classification, clustering, summarization, and question answering~\cite{recent-advancement-embeddings2024, surveryLLM}. Because of their ability to model complex relationships in unstructured text, LLMs are increasingly explored for automating analytical tasks in humanitarian contexts, such as crisis-event classification and early-warning signal detection, rather than merely producing narrative outputs~\cite{surveryLLM}. Nonetheless, it remains unclear whether these models can generate \textit{verifiable, domain-grounded analytical outputs} that meet operational standards in humanitarian work.

Simply prompting LLMs to generate situation reports is often insufficient: models lack access to real-time or specialized data, may produce unverifiable or inconsistent statements, and offer limited control over output structure. Although several approaches exist for automating situation report generation, most lack a clear, easy-to-read structure that enables analysts to quickly grasp essential information. Many systems produce generic summaries~\cite{abdi2017query, cotSummarizer} without mechanisms to assess or improve their quality. SmartBook~\cite{reddy2023smartbook} represents a notable step toward structured report generation, but it remains limited to news-like sources, omits features crucial for situational reporting (such as multi-level summaries and topic-level organization), and does not incorporate internal quality-evaluation mechanisms. These limitations are especially consequential in humanitarian contexts, where accurate and transparently sourced information is essential for decision-making affecting vulnerable populations.

Beyond the humanitarian sector, related AI-driven frameworks are being developed in areas such as climate monitoring and public health, where automated synthesis of complex, multi-source data supports rapid situational awareness~\cite{thulke2401climategpt, ceresa2025RAG-Healt, surveryLLM}. However, few of these systems systematically evaluate generated content against human expert assessments or integrate explicit quality-control mechanisms. Building upon these advances, we propose and evaluate a modular framework for automated humanitarian report generation that emulates human analytical workflows through semantic clustering, retrieval-augmented question answering, and structured summarization. Our system extends prior designs such as SmartBook by incorporating \textit{automatic internal evaluation metrics}, including domain-specific quality assessment~\cite{2022surveyMetricsNLG,2024surveyLLMJudgesGU,2024surveryLLMJudgesLI}, and by introducing visualization modules that contextualize results in terms of the Sustainable Development Goals (SDGs), a widely used framework in humanitarian decision-making~\cite{SDGs-2016, UN_SDG_Goals}. The pipeline is designed to mirror the natural workflow of human analysts: segmenting and clustering content into subtopics, generating targeted questions, and producing evidence-backed answers with citations to source material~\cite{2024questionsevaluations_Survey, reddy2023smartbook, automaticQuestionsLLM}.

To ensure external validity and operational relevance, we collaborated with five experts from the United Nations Children’s Fund (UNICEF) and Data Friendly Space (DFS), organizations that actively develop AI-powered tools for humanitarian analysis. These experts provided feedback throughout system design and participated in evaluating both intermediate components and final outputs. Comparative assessments against alternative systems show that humanitarian professionals consistently judge our framework to produce higher-quality, more actionable, and easier-to-navigate reports. Furthermore, we investigate the use of LLMs as automated judges in humanitarian NLP tasks, comparing their evaluations with expert assessments. Results show strong alignment across multiple dimensions, underscoring the potential of LLMs as reliable evaluators in this domain~\cite{bavaresco2024llms-LLMasJudges,2024surveyLLMJudgesGU}.

This work makes five primary contributions that advance the state of automated humanitarian information analysis:
(i)~we develop a generalizable pipeline applicable to any humanitarian crisis, natural or man-made, capable of handling diverse document types;
(ii)~we introduce a fully automatic, end-to-end reporting system that operates without human intervention, from document ingestion to structured report generation;
(iii)~we conduct extensive evaluation of each pipeline component with humanitarian experts and demonstrate strong correspondence between human and LLM-based quality assessments;
%\ID{(iv)~we release a new dataset of annotated humanitarian questions and answers to foster future research;} 
and
(iv)~we publicly release all code and prompts used to construct our framework, ensuring transparency, reproducibility, and community-driven improvement.

\section*{Methods}

The goal of our framework is to transform raw humanitarian documents into structured, event-specific situational reports through a combination of semantic clustering, question answering, summarization, and visualization components. The methods described in this section detail the dataset used in the study, the design and implementation of the pipeline, and the evaluation procedures adopted to assess the quality and reliability of the generated outputs.  

\subsection*{Dataset}

The development of the pipeline and its evaluation are conducted on a dataset constructed from a large corpus of humanitarian documents provided by our collaborators at Data Friendly Space (DFS) under an established Memorandum of Understanding. The corpus compiled by DFS contains content from 142 distinct online sources, with approximately 40\% originating from \href{https://reliefweb.int}{\textit{ReliefWeb}}, the leading humanitarian information service of the United Nations Office for the Coordination of Humanitarian Affairs (OCHA). In addition to ReliefWeb, the dataset includes material from other humanitarian and disaster-related platforms such as \href{https://thenewhumanitarian.org}{\textit{The New Humanitarian}} and \href{https://www.preventionweb.net}{\textit{PreventionWeb}}, as well as a broad range of global and regional news outlets, including \href{https://www.theguardian.com/}{\textit{The Guardian}},  \href{https://allafrica.com/}{\textit{AllAfrica}}, \href{https://www.aljazeera.com/}{\textit{Al Jazeera}}, and \href{https://www.bbc.com/news}{\textit{BBC News}}.
%Although these documents originate from publicly accessible websites, their redistribution is restricted due to third-party licensing agreements and copyright protections governing the original content. 

As case studies for the situational reports, we define an \emph{event} as an occurrence anchored to a specific geographic area and temporal  (1 week). Event-specific subsets of documents are created by filtering the corpus along these two dimensions. For each document, the referenced country is automatically inferred using a domain-specific geolocation tool~\cite{belliardo2023GeoLocation} and subsequently validated through manual inspection. All documents whose publication date and associated location fall within the spatio-temporal scope of a given event are included in its corresponding subset.
The complete list of events, along with the corresponding number of documents for each, is presented in Table~\ref{tab-events}. The number of documents per event ranges from 47 to 143, resulting in a final dataset of 1,117 documents.

\subsection*{Pipeline}
Starting from the collected documents, our pipeline proceeds in six main stages:

\begin{itemize}
    \item \textbf{Semantic text clustering:} the documents are first segmented into smaller text units and grouped into clusters using HDBSCAN~\cite{hdbscan}. This allows semantically related content to be analyzed together.
    
    \item \textbf{Question generation:} to facilitate exploration of the clustered content, we automatically generate questions using LLMs~\cite{reddy2023smartbook,automaticQuestionsLLM}.
    
    \item \textbf{Answers extraction:} information is automatically extracted from the clustered text segments through a Retrieval-Augmented Generation (RAG) pipeline~\cite{ragOriginalPaper}.

    \item \textbf{Summary generation:}  we generate (i) cluster-level summaries that synthesize all answers associated with each cluster, and (ii) SDG summaries that aggregate answers across clusters for all questions mapped to the same Sustainable Development Goal (SDG) \cite{SDGs-2016}. 
    \item \textbf{Executive summary generation:} in parallel with the QA process, a concise summary of the situation is produced to provide readers with a clear and accessible overview at the beginning of the report.
    \item \textbf{Report Visualization:} Finally, we allow for four modes of visualisation, to facilitate the end users in operationalising the output of the framework, tailoring it to their specific needs.
\end{itemize}

\begin{figure}[htbp]
    \centering
    \includegraphics[width=1\linewidth]{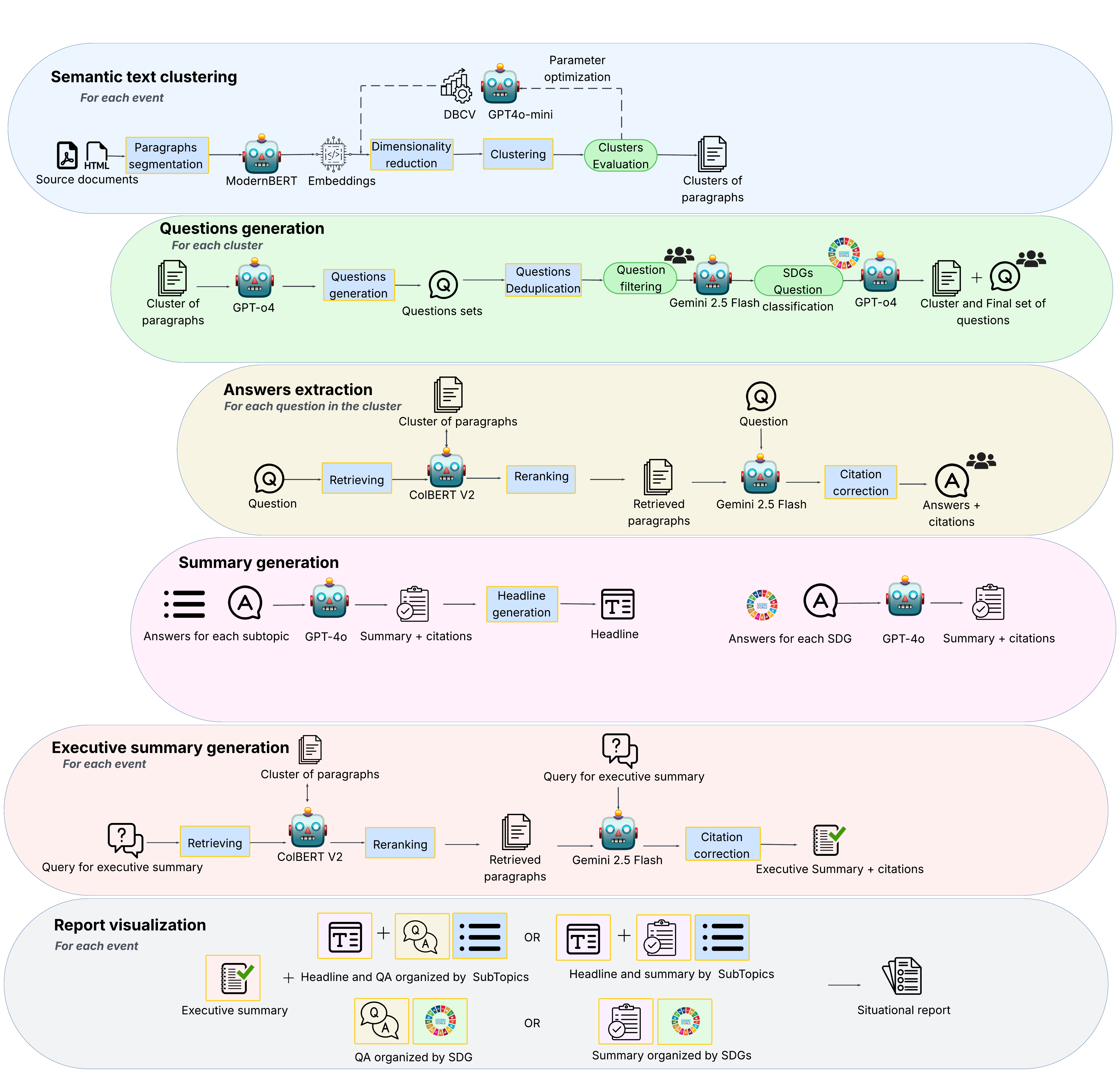}
    \caption{Overview of the proposed pipeline for transforming humanitarian documents into structured situational reports. The process comprises five main phases: semantic text clustering, automatic question generation, automatic answer extraction, creation of the summaries for each topic and for each SDG, executive summary generation,followed by a visualization module enabling four complementary modes of report exploration. Green boxes indicate components evaluated through automated metrics, while icons depicting human figures denote stages assessed by domain experts. }
    \label{fig:pipeline}
\end{figure}

An overview of this pipeline is presented in Figure \ref{fig:pipeline}. The following sections describe each stage in detail.

\subsubsection*{Semantic Text Clustering}  % Semantic Clusters
Initially, we focus on organizing the collected documents into meaningful thematic groups via semantic clustering.

\textbf{Text Preprocessing.} Unlike news articles, general humanitarian documents present unique challenges: they are often lengthy, hierarchically structured, and cover multiple interrelated topics, which limits the effectiveness of conventional document-level clustering approaches such as the one adopted in previous works\cite{reddy2023smartbook}. To address this, we first segment documents into smaller sections prior to clustering. 
This process introduces the well-known challenges of short-text clustering, including data sparsity, high dimensionality, and limited contextual information~\cite{shortText-clustering-review, dimReduction-shortext-hdbscanUmap}. 
After splitting the documents into paragraphs composed of four sentences each to ensure consistent and manageable text segments, 
we perform a series of standard steps~\cite{shortText-clustering-review} to clean and normalize the text:  
we replace encoded characters like smart quotes, dashes, and non-breaking spaces with standard ASCII equivalents and remove unwanted elements such as bullet points, tabs, and URLs while standardizing punctuation.

\textbf{Embedding Generation.}
We embed the normalized paragraphs into dense vector representations employing the 768-dimensional embeddings produced by \emph{ModernBERT}\cite{modernBERT}, a more efficient and higher-performing transformer encoder than the standard BERT model. 
In particular, we use the \texttt{nomic-ai/modernbert-embed-base} model \cite{nussbaum2024nomicembd} implemented in HuggingFace's Transformers library \cite{huggingFaceLibrary}, leveraging its speed and strong semantic representation capabilities. Although LLMs can produce high-quality embeddings \cite{recent-advancement-embeddings2024}, ModernBERT offers a efficient and robust alternative, providing stable semantic representations at a fraction of the computational cost and it has a good placement in the MTEB Benchmark \cite{2022mteb-benchmark-embeddings}. 

\textbf{Dimensionality Reduction.}
High-dimensional embeddings often contain redundancies or noise that can hinder the clustering process, particularly when working with short texts \cite{dimReduction-shortext-hdbscanUmap}. 
To mitigate this issue~\cite{dimreduction2019}, we employ UMAP (Uniform Manifold Approximation and Projection) \cite{umap2018} to decrease the number of dimensions to $k=10$ (with $k$ chosen after manual exploration). 
UMAP is often preferred over methods like PCA and t-SNE due to its balance of speed, scalability, and ability to preserve both global and local data structure \cite{improvingclusering-UMAP}.

\textbf{Clustering.}
For the clustering phase, we employ the Hierarchical Density-Based Spatial Clustering of Applications with Noise (HDBSCAN) algorithm \cite{hdbscan}. 
HDBSCAN  offers several advantages essential for our task. ~\cite{clusteringReview2017}
First, it automatically determines the number of clusters, which is particularly suitable when the number of distinct underlying sub-topics is unknown a priori. 
Further, it effectively identifies clusters of varying densities and constructs a cluster hierarchy, providing richer insights than a single flat partition. 
Its inherent ability to handle noisy data and robustness with minimal parameter tuning make HDBSCAN a well-suited choice for grouping paragraph embeddings.

\textbf{Hyperparameter Search.} 
The quality and interpretability of clustering results critically depend on hyperparameters in both the dimensionality reduction and clustering stages. To navigate this space, we employ a random search strategy, exploring combinations of UMAP's \texttt{n\_neighbors} and \texttt{min\_dist}, along with HDBSCAN's \texttt{min\_cluster\_size} and \texttt{min\_samples}. 
Parameter selection is guided by automatic evaluation metrics. Specifically, we combine two complementary measures with a shared scoring range $[-1, 1]$. 
First, we adopt the Density-Based Clustering Validation (DBCV) index~\cite{dbcv-metric}, specifically designed to assess the quality of density-based cluster structures~\cite{clusteringReview2017}. 
Second, we introduce an LLM-based score, where \emph{GPT-4o-mini} evaluates each cluster according to (i) \textit{semantic coherence} and (ii) \textit{homogeneity}~\cite{2024surveryLLMJudgesLI, 2024surveyLLMJudgesGU}, returning an overall score in the same range as DBCV. 
We choose to employ LLMs as judges since a growing body of evidence shows that they perform reliably in evaluation tasks \cite{thakur2024judging-LLMasJudges,bavaresco2024llms-LLMasJudges}. 
The prompt used for this LLM-based evaluation is reported in the  Supplementary Material S.0.1.1. 
The final hyperparameter configuration is selected by maximizing the mean of the DBCV and LLM-based scores, ensuring clusters are both structurally valid and semantically coherent~\cite{vysala2020evaluating}.

\subsubsection*{Question Generation}
Automatic Question Generation (AQG)~\cite{2024questionsevaluations_Survey} has emerged as a valuable, fast, and cost-effective approach for automatically identifying and extracting the most pertinent information from large sets of documents, regardless of the topic\cite{automaticQuestionsLLM}. Instead of relying on manually defined or static questions, our system dynamically identifies those that are most relevant for situational analysis as the context evolves. We use this method to automatically detect crucial events, trends, and significant developments within each text cluster we previously generate. The aim is to present this key information in the form of questions, with their answers providing a concise summary.
However, AQG alone does not guarantee that these questions are suitable for a situational report, particularly regarding their relevance to the topic and their appropriateness for the humanitarian sector~\cite{automaticQuestionsLLM}. To address this, we've developed a pipeline that filters and refines the generated questions, ensuring they meet the specific needs of a humanitarian report.

\textbf{Prompt Design.} 
To inform the design of prompts for generating questions suitable for humanitarian situational reports, we first conduct consultations with humanitarian experts. 
Experts emphasized that suitable questions should be non-political, focus on urgent, time-sensitive information rather than long-term implications, and be precise and concrete.
Leveraging these insights, we develop three distinct prompts to investigate the impact of prompt design on question generation quality:
\begin{itemize}
    \item \textit{Simple Prompt}: A simple prompt with fundamental instructions and rules designed to filter out poorly formulated questions.
    \item \textit{Chain of Thoughts (CoT)}: Builds on the Simple Prompt by integrating a Chain-of-Thought (CoT) process~\cite{wei2022cot} to guide the LLM towards more relevant questions through step-by-step reasoning.
    \item \textit{CoT + Few Shot}: Extends Chain of Thoughts (CoT) by including examples to facilitate in-context learning, providing demonstrations of the desired question format and content. \cite{brown2020fewshot}
\end{itemize}
Based on our manual evaluation, all prompts performed similarly, but the Simple Prompt was selected because it yielded slightly better results and the added complexity of the other prompts offered no clear benefit for this straightforward task. The prompt can be found in the Supplementary Material S.0.0.1

\textbf{Question Generation.} 
Questions are generated by prompting \emph{GPT-4o} ~\cite{gpt4o-systemcard} (as of February 2025), leveraging the established capability of LLMs to produce relevant questions on specific topics~\cite{automaticQuestionsLLM}. 
The clustered paragraphs are provided in the prompt to ensure contextual grounding. 
To promote diversity\cite{importanceOfDifferentQuestions}, we employ nucleus sampling~\cite{nucleusSampling} and prompt the model three times, producing three possibly redundant sets of questions.  
To remove redundancy, we use the RoBERTa-large model~\cite{roberta} (\texttt{cross-encoder/quora-roberta-base}), pre-trained on the Quora Duplicate Question Pairs dataset, which assigns a score between 0 and 1 indicating the likelihood of two questions being duplicates.  
Redundant questions are filtered out at the generation stage, and a final sample of six questions is taken.

\textbf{Question Filtering.}  
Even with explicit instructions, some generated questions may be unrelated to the main event due to hallucinations or irrelevant references (e.g., mentions of neighboring regions).  
To address this, we develop an automated filtering system using \textit{Gemini 2.5 Flash} \cite{gemini2.5} as an LLM judge\cite{judgingLLMasJudge-Survey}, guided by four binary filtering criteria based on expert feedback:
%\begin{itemize}
 %   \item 
 \textit{a) Not Specific to {country}}: References to locations unrelated to the primary focus.
 %   \item 
 \textit{b)Too Political}: Focus on political causes, opinions, or blame instead of humanitarian impact.
 %   \item 
 \textit{c)Long-term/Historical}: Emphasis on past events or future scenarios lacking current humanitarian relevance.
  %  \item 
\textit{d)Vague/Irrelevant}: Overly broad, abstract, or excessively specific questions.
%\end{itemize}
A question passes the filter only if it receives a 1 for all four criteria.
To reduce the so-called `Self-Prevalence Bias' that can occur when an LLM judges its own output, which is a concern for models such as GPT-4o~\cite{selfPreferenceBias-LLMasJudge}, we use a different model (\textit{Gemini 2.5 Flash}) to evaluate and filter the quality of the answers. We assess the reliability of our judgments against human judgments, and the detailed results can be found in the Results section. 
The optimized prompt for the evaluation can be found the Supplementary Material. (see ~ S.01.2).

\textbf{Question Classification by SDGs.}  
Beyond filtering, each retained question is categorized according to the United Nations Sustainable Development Goals (SDGs) \cite{UN_SDG_Goals, SDGs-2016}.  
We provide the model (again \emph{GPT-4o}) with concise descriptions of all 17 SDGs (see S.0.14 ) and ask it to assign each question to one or more relevant goals.  
The prompt used for this classification is reported in Supplementary Material S.0.13 .
We adopt the SDGs as a classification framework because they are widely recognized in humanitarian contexts as a shared reference point across organizations and stakeholders \cite{SDGs-2016}.  
Moreover, they offer a structured way to group questions that point in the same \textit{direction},for instance, those related to health, education, or climate resilience,thereby enhancing thematic coherence and interpretability in the reports.

\subsubsection*{Answer Extraction}
Standard Large Language Models are prone to generating factual inaccuracies or ``hallucinations'', particularly when dealing with information absent from their training data \cite{ragOriginalPaper,li2025enhancingRAG}. Retrieval-Augmented Generation (RAG) addresses this by dynamically incorporating external knowledge \cite{surveyRAG2024}. The core RAG process involves retrieving relevant documents or passages from an external knowledge base in response to a query, and then providing this retrieved information as context to the LLM during the answer generation phase. This mechanism offers crucial advantages for tasks requiring accurate answer extraction. By grounding the LLM's generation process in specific, retrieved evidence, RAG significantly enhances the factual accuracy and relevance of the outputs, reducing the likelihood of fabricated information \cite{thulke2401climategpt, surveyRAG2024}. Additionally, RAG exhibits flexibility, proving effective not just for general knowledge-intensive tasks like Open-domain Question Answering (OpenQA), but also for handling diverse question types requiring specific contextual information \cite{li2025enhancingRAG}. 
To further enhance trustworthiness and transparency, RAG systems often incorporate citations or attribution \cite{qian2024capacityCitations}, explicitly linking generated statements back to the supporting evidence within the retrieved source documents, which is crucial for enabling factuality verification and boosting model credibility \cite{attributeGenerateRAG}.

\textbf{Paragraph Retrieval.} Given these advantages, %and the requirement in our work to extract precise answers grounded in specific external documents, 
we employ a RAG pipeline using ColBERTv2 \cite{colbertOriginalPaper} for indexing and retrieving paragraphs from our document database. 
ColBERTv2 employs a late interaction mechanism that enables efficient token-level matching between queries and documents and despite being introduced in 2020, ColBERT remains a widely adopted and efficient dense retrieval model \cite{surveyRAG2024}.

\textbf{Reranking and Generation.} To enhance retrieval quality, we implemented RAG-Fusion \cite{ragFusion}, which generates multiple query variations and applies reciprocal rank fusion for robust reranking \cite {li2025enhancingRAG}. The reranked documents, together with the original query and tailored instructions, are passed to the \emph{Gemini 2.5 Flash} model to synthesize answers strictly based on the retrieved context, with instructions to include citations referencing the source documents for each piece of the answer. 
We selected this model as it is freely accessible and yielded superior answer quality in preliminary tests, comparing to \emph{GPT-4o}. 
This model is selected due to its strong performance across various tasks and its favourable placement in the chatbot arena \cite{chiang2024chatbot-arena}.
The prompt used is available in the Supplementary Material S0.1.5
\textbf{Post-processing Citations.} To address instances of incorrect citations, we developed a post-processing correction method. While many studies have demonstrated the effectiveness of such approaches, they often rely on fine-tuned models, which is not feasible in our setting due to limited resources and time constraints \cite{qian2024capacityCitations}. Instead, we drew inspiration from a recent work proposing several lightweight and efficient post-processing techniques \cite{2025citefix}. Among these, a `keyword and semantic context-based matching' approach proved most effective.
This method corrects citations by re-evaluating the relevance of retrieved documents to each claim in the LLM's response. The answer is segmented into discrete claims ($x_i$), each identified by their associated citations using regular expressions to detect citations at the end of factual statements.
For each claim ($x_i$), the goal is to identify the most relevant retrieved document ($\hat{x}_j$) by employing a similarity metric ($s_{ij}$) that must exceed a predetermined threshold. The similarity is computed as:
$s_{ij} = \lambda \cdot J(x_i, \hat{x}_j) + (1 - \lambda) \cdot \text{cosine\_similarity}(q, \hat{x}_j) $, where $J(x_i, \hat{x}_j)$ denotes the Jaccard similarity between the claim and the retrieved document, measuring keyword overlap, and $\text{cosine\_similarity}(q, \hat{x}_j)$ is the cosine similarity between vector embeddings of the original query $q$ and the document $\hat{x}_j$, capturing semantic relevance. We empirically set $\lambda = 0.8$, emphasizing direct keyword matching while incorporating broader semantic context. 
The pipeline is implemented using LangChain to facilitate seamless model integration and customization. 

\subsubsection*{Summary Generation}
To support multi-level analysis of the extracted information, the system produces two different types of summaries. First, for each cluster, we generate a consolidated summary that synthesizes all answers associated with that sub-topic. Second, for each SDG, we generate thematic summaries that aggregate answers across clusters for all questions mapped to that SDG. During summarization, the large language model \textit{GPT4o} is explicitly instructed to preserve all citations associated with each claim, and the full prompt is provided in the Supplementary Material. 
The summary is inputted to \textit{GPT4o} which provides with a headline that describe the topic.

\subsubsection*{Executive Summary Generation}
The final step of our pipeline consists of generating an executive summary that provides a concise, high-level overview of the main findings. 
This summary is constructed exclusively from the paragraphs used to generate the Q\&A answers, ensuring that it remains grounded in the same evidence base. 
To maintain transparency and credibility, citations are preserved and linked back to the original source documents. 
For this purpose, we implement a retrieval-augmented generation (RAG) pipeline, similar to the one described in the previous section, but applied only to the subset of paragraphs used in the answers rather than to the full corpus. 
The specific prompt used for the executive summary generation is provided in the Supplementary Material S0.1.6

\subsubsection*{Report Visualization}
\begin{figure}[!h]
    \centering
    \includegraphics[width=1\linewidth]{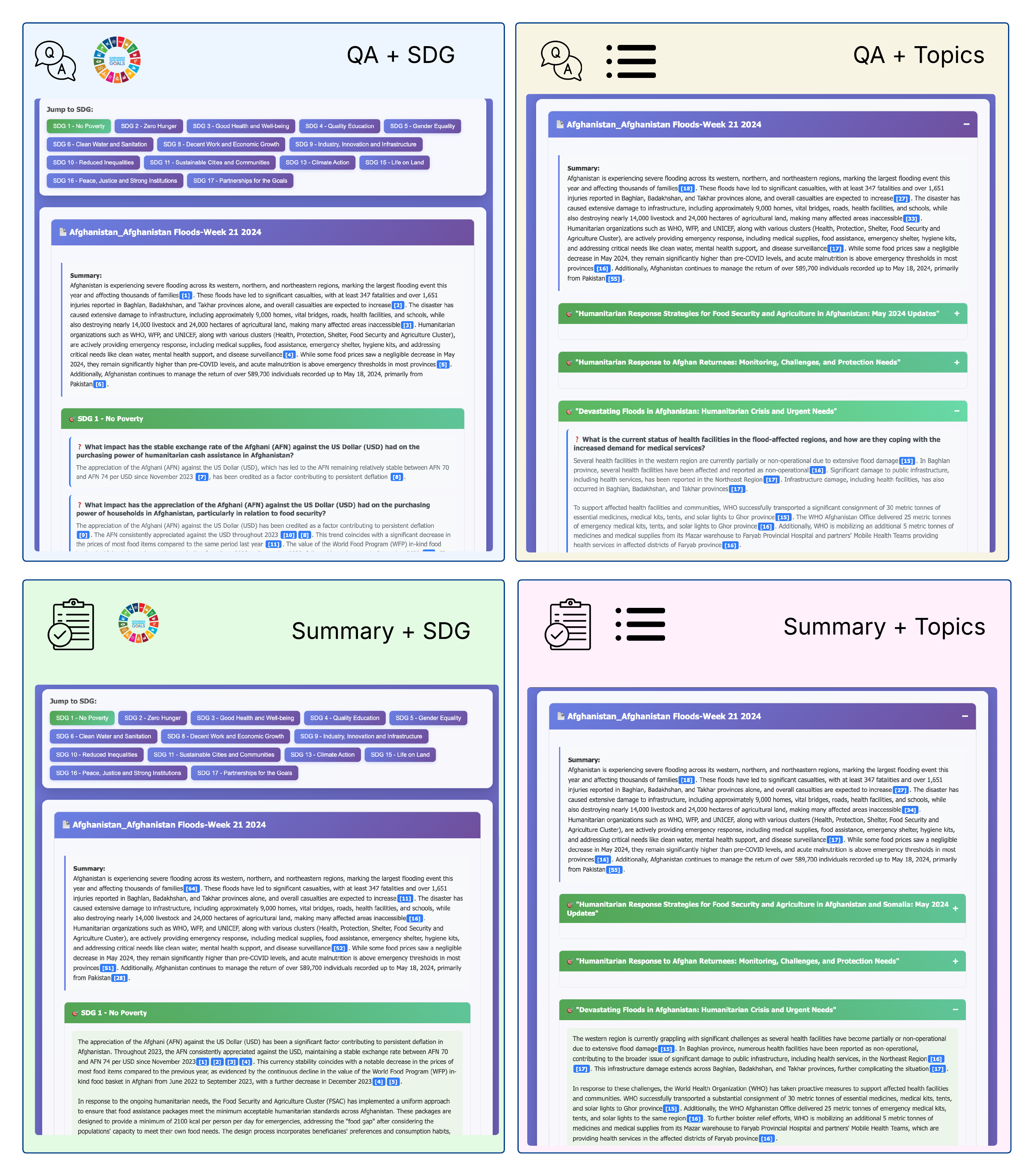}
   \caption{This figure illustrates that our pipeline produces four different versions of the situational report, allowing the user to choose which one to view. The reports can be visualized either organized by Sustainable Development Goals (SDGs) or by automatically discovered sub-topics. Each version can be displayed in the form of a Q\&A or as summarized text. All four versions include an executive summary at the beginning and provide citations for the specific claims made. Each citation includes the original paragraph, the document title, and a link to the source document.} 
   \label{fig:reportsvisualization}
\end{figure}

To enhance readability and usability, the final report is presented through four complementary visualization formats. Specifically, we organize and display the question–answer outputs in the following ways:
(1) grouping questions and answers by sub-topic;
(2) grouping questions and answers by SDGs;
(3) presenting cluster-level summaries; and
(4) presenting SDG-level summaries.
These visualizations allow users to explore the content according to their analytical needs,either by examining detailed, fine-grained responses or by consulting higher-level thematic summaries. The overall structure follows established humanitarian reporting practices~\cite{internews_manual, reddy2023smartbook}, and feedback from senior humanitarian experts at UNICEF and Data Friendly Space (DFS) informed refinements to ensure that the organization supports efficient expert navigation. Finally, all citations are reindexed to ensure unique numbering and consistency across the entire report, merging duplicates and maintaining coherence across all visualization formats and the executive summary. An example of the four report formats is shown in Figure~\ref{fig:reportsvisualization}. 
The reports for the various events can be visualized at this \href{https://decostanzi.github.io/LLM-SituationalReport/Viewer/viewer_v2.html}{link}.

\subsection*{Evaluation}

We thoroughly assess every step of the pipeline, focusing on the quality of the generated questions, the quality of the answers and their associated citations, and the overall quality of the final report. To ensure a comprehensive assessment, these evaluations were performed not only by the authors but also by human participants, including humanitarian experts such as personnel from UNICEF and independent researchers. This multi-perspective approach allows us to capture both domain-specific insights and unbiased external judgments. 
The study received ethical approval from the ISI Foundation Ethical Review Board (IRB) and was conducted in compliance with institutional policies and applicable data protection regulations. All evaluators provided informed consent prior to participating in the study. No sensitive personal data were collected or processed, and expert evaluators participated voluntarily after being fully informed about the study procedures. \\
The full set of results are available upon request, while additional details about the expert participants and the evaluation guidelines can be found in the Supplementary Information. No sensitive personal data were collected beyond professional feedback.

\textbf{Evaluation of Question Generation and Filtering.}
We evaluate the automatic question generation process on three aspects:
(i) prompt optimization,
(ii) the effectiveness of the filtering mechanism, and
(iii) the quality of the final questions after filtering.
To find the best prompt for question generation, we manually examine the outputs of the three prompt designs (Prompt~1–3) to assess their suitability for humanitarian situational reports, using the criteria of \emph{Relevance} and \emph{Importance}. %, and \emph{Urgency}. % 
Here, \emph{Relevance} (standard measure in Information Retrieval \cite{definitionRelevance}) reflects whether the question pertains directly to the specific humanitarian event, and
\emph{Importance} captures whether the question is valuable for inclusion in a humanitarian situation report.
We are especially interested in the latter measure of the kind of information that is most useful to humanitarian efforts, beyond simple relevance. 
To ensure an unbiased evaluation, we randomly select 33 questions from the topics generated by each prompt (totalling 100 questions overall) and assess their quality without knowing which prompt had produced them.
The evaluation is conducted independently by the three authors, with partial overlap in their annotations to allow for consistency checking and inter-annotator agreement assessment.

For the filtering of the generated questions, we assess GPT-4o’s performance on binary classification tasks concerning the four metrics outlined in the Question Filtering subsection against human annotations that are performed by the authors. We summarize its performance by computing precision, recall, and F1-score. This evaluation follows established practices for assessing LLMs as judges~\cite{2024surveyLLMJudgesGU, 2024surveryLLMJudgesLI}. In this phase, 200 questions were randomly sampled from the set of generated events and annotated according to the four metrics, again involving the three authors with overlapping assignments to ensure consistency and reliability. 

The quality of the final set of generated questions, obtained after filtering and using the optimal prompt strategy, is assessed by three humanitarian experts according to three binary criteria: \emph{Relevance} and \emph{Importance}, as defined above, as well as 
\emph{Urgency}, which captures whether the question addresses short-term needs such as food, water, shelter, health, or protection. 
In total, 150 questions were evaluated in this phase, with some overlap across annotators to support consistency assessment.
The \emph{Urgency} dimension is introduced by the humanitarian experts during the exploratory feedback session to better capture immediate humanitarian priorities. Information about the annotators can be found in the Supplementary Material S4. 

We compute inter-annotator reliability to measure the consistency of judgments both among human annotators and, where appropriate, between human annotators and the LLM output. Specifically, we calculate the Percentage of Agreement, Cohen’s Kappa~\cite{cohenkappa}, and the Prevalence-Adjusted Bias-Adjusted Kappa (PABAK), which is less sensitive to the prevalence paradox - the phenomenon where Kappa can appear artificially low when the distribution of categories is highly imbalanced, even if agreement between annotators is high. ~\cite{limitationsKappa, limitationskappa2}. The same inter-annotator reliability analysis is conducted for the evaluation of answers (see the next Section).
The detailed annotation guidelines for each of the above tasks can be found in the Supplementary Material S0.2.

\textbf{Evaluation of Generated Answers.}
\label{sec:answersEvaluationMethod}
To evaluate the quality of the generated answers for the questions that passed our filtering system, we conduct an extensive manual annotation effort involving 12 annotators familiar with the task and humanitarian domain.  
The evaluation focus on two primary metrics: \textit{Answer Relevance}~\cite{2024ragas-rag} and \textit{Citation Quality}~\cite{qian2024capacityCitations, gao2023-citations}.  
We randomly select 10 question–answer pairs from each of the 13 humanitarian events, resulting in a total of 130 evaluated answers. These are distributed among the 12 annotators with partial overlap to allow for consistency checking and inter-annotator reliability assessment.  
Across this sample, annotators evaluate 983 individual claims and 1,146 supporting citations.  
To ensure the robustness of the evaluation, we compute inter-annotator agreement to verify the consistency of judgments among annotators.

We assess \textit{Answer Relevance} by determining whether the generated answer provided information directly pertinent to the posed question. A binary label was used: relevant or not relevant~\cite{2024ragas-rag}.  
\textit{Citation Quality} is evaluated through \textit{Citation Recall} and \textit{Citation Precision}, following definitions from prior work~\cite{gao2023-citations, qian2024capacityCitations}.  
Specifically, for each answer, composed of $n$ statements $s_1, \ldots, s_n$, each statement $s_i$ cites a set of passages $C_i = \{c_{i,1}, c_{i,2}, \ldots\}$.

\begin{itemize}
    \item \textit{Citation recall} evaluates whether the entire statement $s_i$ is supported by its associated set of cited passages $C_i$.
    \item \textit{Citation precision} measures whether each individual citation in $C_i$ is relevant to the statement $s_i$.
\end{itemize}

To operationalize this for human annotators, a three-point scale is adopted for both citation recall and precision:
\begin{itemize}
    \item \textit{Fully supports}: The citation (for precision) or the set of citations (for recall) entirely substantiates the statement $s_i$.
    \item \textit{Partially supports}: The citation or set of citations offers partial but incomplete support for $s_i$.
    \item \textit{Does not support}: No support for $s_i$ is provided.
\end{itemize}
When a single citation is associated with a single statement, citation recall and precision are equivalent.

\textbf{Usability Evaluation.} After generating the final report, we assess its quality through a structured evaluation involving a panel of 5 humanitarian experts. The evaluation consist of three stages. In the first stage, experts examine the individual components of our system, providing feedback on their quality and their usefulness. In the second stage, they compare outputs from three systems,our proposed model, Smartbook \cite{reddy2023smartbook}, and a state-of-the-art multi-document summarization system based on large-context prompting \cite{cotSummarizer},without being informed which system produced each output. Finally, in the third stage, the experts rank the systems overall and provid open-ended qualitative comments regarding their perceived strengths and limitations.
The survey is designed following a five-point Likert scale adapted from prior work \cite{reddy2023smartbook} to capture quantitative assessments of quality, coherence, and informativeness. To ensure a fair comparison, we standardize the underlying model architecture across systems, employing Gemini 2.5 Flash \cite{gemini2.5} for both Smartbook and the reference summarization baseline. This approach ensure consistent technological conditions across all experiments.
The detailed survey instructions and question items are provided in Supplementary material S0.2.4.

\section*{Results}
In summary, for a given event defined by a specific country and time period, our pipeline segments and clusters the document texts, generates questions with corresponding answers and citations, produces summaries at both the cluster and SDG levels and produces an executive summary at the beginning of the report. We then evaluate the system’s performance at each stage of this process.

\subsection*{Semantic Text Clustering} %%% Make sure the titles are the same as in Methods
The first step of the pipeline involves the topical clustering of the text segments from the documents relevant to the event. Table~\ref{tab-clustering-results} reveals significant variation in the number of clusters across events. While some, such as the Afghanistan Floods and Indonesia's events, show a smaller number of clusters (7 each), others, for instance, the May Israel-Hamas war, have a much higher count. This wide range, from 6 to 32 clusters, highlights the diverse structural complexity of the data for each humanitarian crisis and this is not correlated to the amount of documents available (Pearson coefficient = 0.1).  
Interestingly, the combined semantic coherence and homogeneity score produced by the LLM-as-a-judge remains consistently high, generally above 0.83, even with this diverse clustering, suggesting the LLM's robust clustering performance across varying data structures. 
There is a moderate correlation between the Density-Based Clustering Validation (DBCV) and the LLM score, the Pearson coefficient between them is 0.4, sign that there is a relation between the metrics we used.

\begin{table}[h!]
\centering

% ---------------- (a) ----------------
\begin{subtable}{\textwidth}
\centering

\begin{tabular}{lllll}
\toprule
& \textbf{Event} & \textbf{Period} & \textbf{Type} & \textbf{N Docs} \\
\midrule
1  & Floods and volcanic activity in Indonesia & May 06 – May 12 & Natural Disaster  & 61 \\    
2  & Afghanistan Floods & May 13 – May 19 & Natural Disaster  & 67\\
3  & Cyclone Remal in Bangladesh & May 13 – May 19 & Natural Disaster  & 53\\
4  & Hurricane Beryl in Jamaica* & Jul 08 – Jul 14 & Natural Disaster & 47 \\
5  & Landslide and Floods in India & Jul 22 – Jul 28 & Natural Disaster  & 84\\
6  & Flooding in Nigeria & Sep 9 - Sep 15  & Natural Disaster  & 98 \\[4pt]
7  & Ukraine conflict & May 27 – Jun 02 & Conflict/Violence & 61\\
8  & Israel-Hamas war* & May 6 - May 12  & Conflict/Violence & 122 \\
9  & Sudan conflict* & Aug 19 – Aug 25 & Conflict/Violence & 100\\
10 & UK riots & Jul 29 – Aug 04 & Conflict/Violence & 86\\
11 & Sudan conflict & Sep 16 – Sep 22 & Conflict/Violence & 126\\
12 & Gang violence and humanitarian crisis in Haiti & Sep 23 – Sep 29 & Conflict/Violence  & 69\\
13 & Israel-Hamas war & Sep 30 – Oct 06 & Conflict/Violence & 143 \\
\bottomrule
\end{tabular}

\caption{}
\label{tab-events}

\end{subtable}

\vspace{0.8cm}

% ---------------- (b) ----------------
\begin{subtable}{\textwidth}
\centering

\begin{tabular}{lllll}
\toprule
 & \textbf{Event} & \textbf{Number of Clusters} & \textbf{DBCV} & \textbf{LLM Score} \\
\midrule
1  & Floods and volcanic activity in Indonesia & 7 & 0.685 & 0.943 \\    
2  & Afghanistan Floods & 7 & 0.411 & 0.938 \\
3  & Cyclone Remal in Bangladesh & 11 & 0.360 & 0.964 \\
4  & Hurricane Beryl in Jamaica & 9 & 0.415 & 0.984 \\
5  & Landslide and Floods in India & 16 & 0.402 & 0.835 \\
6  & Flooding in Nigeria & 6 & 0.299 & 0.849 \\
7  & Ukraine conflict & 15 & 0.552 & 0.860 \\
8  & Israel-Hamas war & 32 & 0.292 & 0.837 \\
9  & Sudan conflict & 6 & 0.318 & 0.903 \\
10 & UK riots & 7 & 0.251 & 0.864 \\
11 & Sudan conflict & 7 & 0.549 & 0.830 \\
12 & Gang violence and humanitarian crisis in Haiti & 19 & 0.459 & 0.870 \\
13 & Israel-Hamas war & 7 & 0.683 & 0.996 \\
\bottomrule
\end{tabular}
\caption{}

\label{tab-clustering-results}

\end{subtable}

\caption{(a) Events used to test our framework with the period in which they occurred and the number of documents we have available for each event. (b) Clustering performance results for each event showing the number of clusters identified, DBCV (Density-Based Clustering Validation) score, and LLM evaluation score for homogenity and coherence.}
\end{table}

\subsection*{Question Generation}
\textbf{Prompt Design.} We compared three approaches: a simple prompt, a Chain-of-Thought (CoT) prompt, and a CoT prompt enriched with few-shot examples. We manually evaluated a sample of 100 questions,roughly one-third from each prompt,for their relevance to the humanitarian context and overall importance. Inter-annotator agreement was computed using 15 shared questions per annotator, and we observed perfect agreement across all evaluations.
The results show that the simple prompt performs slightly better than the more complex alternatives. It achieves the highest relevance score (85.1\%), compared to 81.2\% for CoT and 79.3\% for CoT with few-shot examples. A similar pattern holds for importance: the simple prompt reaches 96.5\%, outperforming both CoT (93.3\%) and CoT with few-shot examples (93.6\%). 
Given this evidence, we adopt the simple prompt in our final pipeline, described in Supplementary material S.0.1. .

\textbf{Question Filtering.}
To assess the reliability of our Gemini based filtering system, we evaluate a dataset consisting of 200 questions generated before filtering, consisting of two questions for each cluster and for each event.
These questions are then coded for the four filtering metrics: Relevance,Political,Long term, and Generality, as defined in the Methods section.
The inter-annotator agreement, detailed in Supplementary Table S1, demonstrates good overall reliability in our human annotation process. Specifically, the PABAK scores indicate substantial agreement for Relevance (0.867) and Political (0.822). Agreement for Long term (0.778) is also good, while Generality (0.644) show moderate agreement. As anticipated, Cohen's Kappa values were lower across the categories, likely reflecting the impact of the prevalence paradox within our dataset, however, the robust PABAK scores confirm the consistency of our human judgments. The final labels are selected randomly where annotators disagreed.

Using this annotated dataset, we evaluate the performance of the Gemini 2.5 Flash filtering model by comparing its judgments against the aggregated human annotations. 
Table~\ref{tab:agreementLLM} presents the precision, recall, and F1 scores for this comparison. The system demonstrate notably high precision across all individual metrics, for instance, achieving 94.7\% for Relevance and 93.3\% for Political. This suggests that when the LLM identifies a question as meeting a specific criterion, it is highly likely to be correct. Recall scores are also strong, particularly for Generality (87.2\%) and Political (85.9\%), indicating the LLM's effectiveness in identifying most of the instances of these categories. Consequently, the F1 scores, which balance precision and recall, are high for Generality (89.2\%) and Political (89.4\%), and also reflected strong performance for Long term (86.5\%) and Relevance (86.3\%). 
For the overall decision of whether to \textit{``Keep the question''}, the system achieve a balanced F1-score of 79.1\%, underscoring good practical alignment with human consensus. To quantify statistical uncertainty, we computed 95\% confidence intervals (CIs) for precision, recall, and F1-scores across the four filtering metrics using bootstrap resampling ($n = 1{,}000$). The mean F1-score for the overall decision was $79.1\%$ (95\% CI [73.4, 84.3]), while the individual metrics show similarly narrow intervals ranging from $\pm 3.9$ percentage points for \textit{Relevance} to $\pm 3.4$ percentage points for \textit{Generality}. These results confirm that performance differences across criteria are statistically meaningful rather than due to sampling variability.

\begin{table}[h!]
    \centering
    \small
    \begin{tabular}{lcccc}
        \toprule
        \textbf{Category} & \textbf{Agreement (\%)} & \textbf{Precision (\%)} & \textbf{Recall (\%)} & \textbf{F1 (\%)} \\
        \midrule
        Relevance & 77.7 [71.8, 83.2] & 94.7 [91.0, 98.0] & 79.4 [73.3, 85.0] & 86.3 [82.2, 89.9] \\
        Political & 82.1 [76.7, 87.1] & 93.3 [89.1, 97.0] & 85.9 [80.7, 90.3] & 89.4 [85.9, 92.5] \\
        Long term & 78.2 [72.8, 83.7] & 89.8 [85.0, 94.2] & 83.5 [77.9, 88.8] & 86.5 [82.5, 90.2] \\
        Generality & 81.2 [75.7, 86.2] & 91.3 [86.8, 95.2] & 87.2 [82.3, 91.8] & 89.2 [85.6, 92.4] \\
        Keep the question & 74.2 [68.3, 80.2] & 82.4 [75.9, 88.6] & 76.2 [69.0, 83.3] & 79.1 [73.4, 84.3] \\
        \bottomrule
    \end{tabular}
    \caption{Agreement between human and LLM scores for the quality of the generated questions, expressed in percentages with 95\% confidence intervals in brackets.}
    \label{tab:agreementLLM}
\end{table}

\textbf{Question Generation.}
The quality of the final set of generated questions (after using the optimal prompt strategy and filtering) is assessed by three humanitarian experts based on Relevance, Importance, and Urgency.
Considering limited time availability, we select the three topics from the Table~\ref{tab-events} marked with * for this expert evaluation, as they cover two different natural disasters and a conflict.  
The experts evaluate 150 questions, with an overlap of 15 questions between them. 
Inter-annotator agreement for this final quality assessment is presented in Supplementary Table S2. 
The PABAK scores indicate fair to moderate agreement across the three metrics: Relevance (0.595), Importance (0.563), and Urgency (0.510). This suggests that while there is a general consensus, the subjective nature of these qualities can lead to some divergence in expert opinions. As with the filtering evaluation, Cohen's Kappa scores are lower, highlighting the effect of score distribution on this metric.
Considering the performance of the system, Table~\ref{tab:detailed_scores} summarizes the scores provided by each expert for each of the metrics and their mean scores. On average, questions were rated highly for Relevance (84.7\%) and Importance (84.0\%), and moderately for Urgency (76.4\%). Expert 1 and Expert 3 provide consistently higher scores across Relevance and Importance compared to Expert 2, while scores for Urgency showed less variance among experts. These results point to the diversity of opinion possible even among experts. 

Overall, the experts note that the evaluated questions contained a good mix of detailed and broader questions, with the detailed questions being particularly valuable for creating situational reports. They also observed that some topics produce higher-quality questions than others. Questions that are rated as less relevant typically addressed very specific topics mentioned in the source documents but are not essential for the situational report. For example, the question ``What measures are being taken to ensure the safety and legal compliance of Cash for Work participants in Khyber Pakhtunkhwa, particularly concerning occupational safety and health standards?'' refers to a program mentioned in the documents but is not directly relevant to a situational report focused on the floods in Pakistan. These observations highlight the diversity of opinion possible even among experts and point to areas where the system could be further refined. 
\begin{table}[h!]
    \centering
    \begin{tabular}{lcccc}
        \toprule
        \textbf{Metric} & \textbf{Expert 1 (\%)} & \textbf{Expert 2 (\%)} & \textbf{Expert 3 (\%)} & \textbf{Mean Score (\%)} \\
        \midrule
        Relevance & 92.7 & 72.7 & 88.7 & 84.7 \\
        Importance & 94.5 & 67.3 & 90.3 & 84.0 \\
        Urgency & 70.9 & 74.5 & 83.9 & 76.4 \\
        \bottomrule
    \end{tabular}
    \caption{Evaluation scores for the quality of the generated answers by the experts, expressed in percentages.}
    \label{tab:detailed_scores}
\end{table}

\subsection*{Answer Extraction and Citations}
To evaluate the quality of the generated answers, we perform an extensive labelling effort resulting in 983 claims and 1146 citations annotated for 130 questions with the help of 13 independent annotators.
The agreement on Answer Relevance is perfect among all of the annotators, whereas it was at 0.90 for Citation Precision and 0.81 for Citation Recall, as can be seen in Supplementary Table S3. \\
% For metrics where disagreements occurred among annotators, we randomly selected one of the annotated values to represent the final score for that particular instance.
Table ~\ref{tab:final_quality_assessment_answers} summarizes the system's performance.
We find that a large percentage 86,3\% of the answers are deemed relevant. For citation quality, a substantial portion of citations are found to be relevant or partially relevant, both in terms of precision and recall. Specifically, for citation precision, 76,3\% of citations are relevant and 5,93\% are partially relevant, out of 1146 evaluated citations. 
For citation recall, 76,4\% of claims are supported and 5,90\% are partially supported, out of 983 sets of citations evaluated.
This assessment highlights a strong overall performance of the model in both answer relevance and citation quality. 

\begin{table}[h!]
    \centering
    \begin{tabular}{lcccc}
        \toprule
        \textbf{Metric} & \textbf{Count} & \textbf{ Relevant (\%)} & \textbf{Supported (\%)} & \textbf{Partially Supported (\%)} \\
        \midrule
        Answer Relevance    & 130 & 86.3 & -- & -- \\
        Citation Precision  & 1146 & -- & 76.3 & 5.93  \\
        Citation Recall     & 983 & -- & 76.4 & 5.90  \\
        \bottomrule
    \end{tabular}
    \caption{Quality assessment of generated answers and citations. }
    \label{tab:final_quality_assessment_answers}
\end{table}

We also evaluate answer relevance using an LLM as judge \cite{2024ragas-rag}, as we did in the previous section. 
To assess the reliability of this evaluation, we again employed bootstrap resampling ($n = 1{,}000$) to compute 95\% confidence intervals.
The agreement with human annotators was high, with a Percentage of Agreement of 70.1\% (95\% CI [67.1, 74.2]), Precision of 69.5\% (95\% CI [67.3, 72.9]), Recall of 97.1\% (95\% CI [95.3, 98.8]), and F1 Score of 81.0\% (95\% CI [79.6, 83.4]), confirming good overall alignment between human and LLM judgments, consistent with the results reported in Table~\ref{tab:agreementLLM} for the quality of the questions.
Those classified as Not Relevant are typically answers where the model inferred a response even when the sources did not provide sufficient information, often using phrases such as \emph{``The sources do not provide clear answers to that''}, followed by a speculative or tangential response. Another common issue arose when the answer closely mirrored the phrasing of the question without adding substantive information. For example:
\textit{``\textbf{Q:} What specific warnings have been issued to local aviation authorities regarding the volcanic activity of Mount Ibu?''
``\textbf{A:} The country's volcanology agency issued a warning for aviation authorities managing local flights\textsuperscript{[41]}.''}
In such cases, although the answer technically responds to the question, it does so with minimal value-added content, often lacking synthesis or analytical depth.

Turning to citations, we first consider citation recall, which evaluates whether all claims made in the answer are supported by the retrieved evidence. We find that 86.6\% were fully or partially supported. Notably, in all cases where claims were \emph{not} supported (13.37\%), there was only one citation associated with the claim, it's the case in which citation recall is exactly the same as citation precision. 
Regarding citation precision, 86\% of citations were judged as relevant or partially relevant, indicating a generally strong link between statements and their supporting references. However, the remaining 14\% highlight important limitations. The primary sources of citation error are:
(i) \textit{Unnecessary citations}: Additional references included even when not needed.
(ii)\textit{Inferred citations}: The claim is generally supported by the cited sources, but not stated explicitly; it is derived by interpreting or combining information across multiple contexts.
(iii) \textit{Missing citations}: Statements are made without any explicit support in the retrieved documents.
(iv) \textit{Hallucinated content}: Entirely fabricated statements not grounded in any source material.
Among these, hallucinations are particularly problematic, as they represent a breach of factual reliability. This issue has been widely discussed in recent literature on large language model behavior~\cite{attributeGenerateRAG,rag2025NotSafer}, and poses significant challenges for the deployment of such systems in sensitive or high-stakes domains.
A manual analysis of citation errors shows that 5.1\% of citations are classified incorrect, 4.5\% were inferred, 4.3\% as unnecessary, and 3.0\% as hallucinations. 

\subsection*{Overall Report Evaluation}
We conclude by presenting the expert evaluation of the reports produced by our system, in comparison with two others. 
Across all stages of the final report evaluation, a total of five experts with direct experience in producing and interpreting humanitarian situational reports participated in the assessment, including four analysts from UNICEF and one technical specialist from Data Friendly Space.
In the first stage of evaluation, experts examined individual components of our system. 
They expressed strong appreciation for features unique to our approach, particularly the executive summary at the beginning of the reports and the ability to explore the report through multiple structures and visualizations. This  flexibility, absent in SmartBook, was repeatedly highlighted as important for accommodating diverse analytical preferences. 
Experts also emphasized the importance of citation transparency in humanitarian reporting; this requirement is met by both our system and SmartBook, but is entirely missing from the summariser \cite{qian2024capacityCitations}.

\begin{figure}[!h]
    \centering
    \includegraphics[width=1\linewidth]{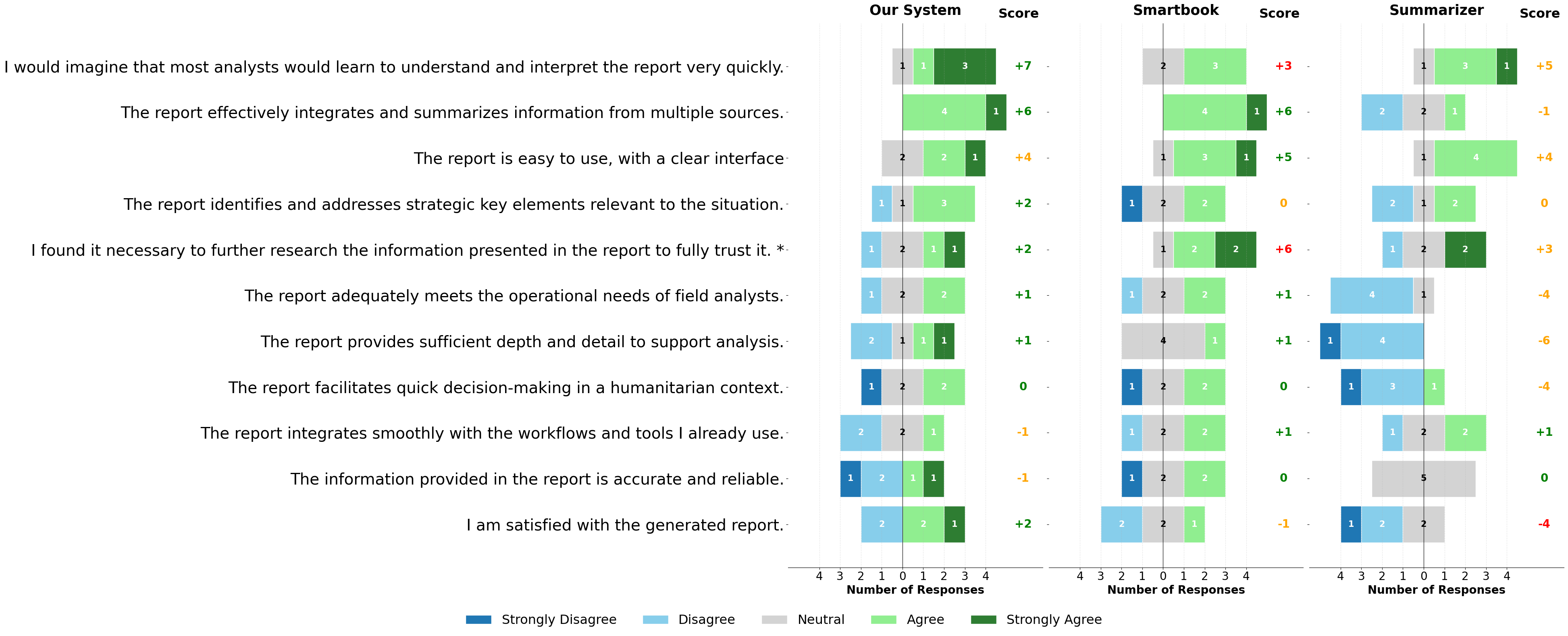}
    \caption{Expert evaluation of the three reporting systems: (i) our proposed framework, (ii) SmartBook, and (iii) a summarizer, across multiple dimensions of report quality. Evaluations were conducted by five humanitarian experts with direct operational experience in producing and interpreting situational reports. Responses are shown on a five-point Likert scale, ranging from ``Strongly Disagree'' (dark red) to ``Strongly Agree'' (dark green). For each criterion, coloured bars represent the number of expert ratings falling into each category. Higher concentrations of green indicate stronger perceived performance. The results show that our system is rated highest for analytical depth, integration of information, and support for humanitarian decision-making, while SmartBook performs strongly on ease of use. The summarizer receives predominantly neutral ratings due to its lack of structure and citations. }
    \label{fig:comparisonsystems}
\end{figure}

In the second stage, experts compared full reports from all three systems using a five-point Likert scale. The quantitative results are shown in
Figure ~\ref{fig:comparisonsystems}.   
Several clear patterns emerge from these evaluations.
First, our system achieves strong performance in dimensions directly related to analytical quality. Experts rated it highly for its ability to identify and address strategic elements relevant to the situation, integrate information from multiple sources, and provide depth sufficient to support operational decision-making. In these categories, our system consistently exhibits the largest proportion of ``Agree'' and ``Strongly Agree'' responses.
SmartBook performs well on ease-of-use and workflow integration, reflected in the higher scores shown in Figure~\ref{fig:comparisonsystems} for these specific dimensions.
The expert evaluation demonstrates that our system performs comparably to SmartBook on several core dimensions of report quality, with both systems sharing a similar overall approach while differing in specific implementation details. 
However, these advantages must be interpreted in light of its TF-IDF–based clustering, which is well suited to news-like corpora but less compatible with heterogeneous humanitarian documents like ours. This approach led to many clusters containing only two documents, which SmartBook discards, resulting in the exclusion of a substantial portion of the corpus and, consequently, the generation of shorter and simpler reports. 
This reduction in coverage explains the high usability ratings but also highlights an important limitation when handling heterogeneous humanitarian text.

Reliability is an area where none of the systems received particularly strong evaluations. This outcome is consistent with the stringent expectations of the humanitarian domain, where even a single incorrect or unsupported citation can undermine trust in the entire report. To contextualize the reliability scores, we evaluated a sample of 150 claims generated by SmartBook; The results closely mirrored those obtained for our system: 81\% of citations were relevant or partially relevant and 84\% of claims were supported or partially supported. Notably, however, SmartBook achieved these metrics on a truncated document set; our RAG and post-processing pipeline likely offers a more reliable solution for accurate citation alignment when processing the full volume of evidence. This suggests that, under equivalent evidence conditions, our approach is likely to provide more robust citation alignment.
The summarizer system, while faster, retains the advantage of grounding all content strictly in the source texts but lacks the organizational depth and analytical features that experts deem essential for complex situational analysis.

In terms of informativeness and coverage, both our system and SmartBook outperform the baseline summarizer, which received consistently neutral ratings owing to its lack of citations and organizational structure.
When asked to provide an overall ranking, three of the four experts selected our system as their preferred choice, citing its greater comprehensiveness and navigability relative to the alternatives. They attributed these advantages to the integration of multiple organizational modes (subtopics and SDGs), granular question–answer structures, and higher-level summaries. These features were viewed as essential for synthesizing complex humanitarian crises.
In the final ranking task, three of the four experts selected our system as their preferred option. The qualitative feedback indicates that while no single visualization mode is universally optimal, the ability to switch between complementary organizational structures, together with comprehensive evidence integration and transparent citation, makes the system particularly well suited for operational use.

Finally, we note that the evaluations were conducted across several distinct humanitarian events. This diversity likely contributed to some variability observed in the scores for information quality and report usefulness, reflecting differences in the quality and density of source materials.
Similar variability was previously observed in the question generation stage and is consistent with differences in document availability and specificity across events.

\subsection*{Computational Efficiency}

To assess the computational efficiency of the proposed framework, we compared the total processing time required to generate situational reports across the three systems on a MacBook Pro (13-inch, 2021) equipped with a 2.3 GHz Quad-Core Intel Core i7 processor and 32 GB of RAM. As expected, the large-context summarizer was the fastest, owing to its comparatively simple architecture, which performs direct multi-document summarization without intermediate analytical steps. 
Processing times for our system and SmartBook were broadly comparable. On average, our pipeline required an additional 2–3 minutes to complete a report. This difference is attributable to components unique to our design, including the generation of executive summaries, question filtering, RAG-based answer extraction with post-processing of citations, and the production of cluster-level and SDG-level summaries. While these steps introduce modest computational overhead, they substantially enhance report depth, structure, and analytical utility.

From an operational cost perspective, only components relying on GPT-4o incur per-execution expenses, while the majority of models employed, such as Gemini 2.5 Flash and ModernBERT, are freely available. Despite the inclusion of GPT-4o, the average cost of generating a full report remains below \$2.00 USD per event. Notably, for several tasks (e.g., question answering), the freely available Gemini 2.5 Flash model not only reduced costs but also produced higher-quality outputs in preliminary evaluations.

Importantly, the modular design of the framework enables flexible substitution of model components, making it straightforward to optimize for speed, cost, or performance as needed. Further runtime reductions could be achieved by batch retrieval strategies, parallelization of cluster-level operations, or integration of smaller, domain-adapted generative models.
Benchmark execution times for three representative events (marked with * in Table~\ref{tab-events} ) are provided in the Supplementary Materials.

\section*{Discussion}

This study demonstrates that large language models can effectively support the automatic generation of humanitarian situation reports, one of the most demanding analytical tasks in crisis response \cite{alnap_ai_humanitarian_2025}. The proposed framework integrates document clustering, automatic question generation, retrieval-augmented answer synthesis, and executive summary generation within a coherent pipeline designed to approximate the reasoning workflow of human analysts. Applied across 13 humanitarian events and more than 1,100 source documents, the system successfully extracted, organized, and synthesized critical information with high factual accuracy and consistency.

Five humanitarian experts with direct experience producing and interpreting situational reports evaluated the system. They highlighted several strengths. The generated questions were consistently rated as highly relevant, important, and urgent, and the corresponding answers were judged to be well grounded in evidence with generally precise citation alignment. Experts from UNICEF further emphasized the system’s operational usefulness and its potential for integration into existing analytical workflows. Compared with existing approaches, most experts preferred our system, citing the flexibility provided by its multiple organizational modes (subtopics and SDGs), the usefulness of executive summaries, and its capacity to structure and contextualize information in a manner aligned with established humanitarian analytical practices. These features collectively enhanced usability and analytical depth relative to baseline systems. Although SmartBook processed some document types more quickly, our system’s additional components, including executive summaries, question filtering, and cluster-level synthesis, yielded reports that were more comprehensive and better adapted to heterogeneous humanitarian text sources.

Despite these strengths, several limitations remain. The system’s outputs exhibit variability that reflects differences in the quality, completeness, and clarity of the underlying humanitarian documents; data gaps and inconsistencies can propagate through later stages, affecting both the specificity of generated questions and the informativeness of synthesized answers. Citation reliability, while generally strong, is not perfect: occasional citation errors persist, and even a single unsupported or incorrect citation can undermine trust in operational settings where precise attribution is essential. More broadly, the system currently processes only textual inputs, excluding potentially valuable information contained in images, tables, and geospatial data. Runtime costs also constrain real-time or large-scale deployment, though the modular architecture facilitates future optimization and substitution with more efficient or domain-adapted models.

Future work should address these limitations by extending the framework to multimodal and multilingual inputs, leveraging vision–language models to incorporate visual and structured evidence, thereby capturing a richer representation of humanitarian conditions. Additional factuality metrics, such as FactScore \cite{min2023factscore}, may further strengthen robustness, while systematic evaluation across general, specific, and quantitative question types could clarify the system’s strengths and failure modes. Embedding human feedback throughout the pipeline will remain essential for ensuring oversight and accountability as such systems move toward operational integration. Experts also recommended enabling more granular control over report length, beyond fixed presets, and incorporating additional visualization features, including timelines, severity indicators, and cluster-level summaries, to support rapid situational understanding. Linking the framework with existing platforms such as HDX Signal could enhance real-time monitoring and longitudinal crisis analysis.

A complementary finding of this study concerns the emerging role of LLMs as evaluators. The strong correspondence observed between human judgments and LLM-based assessments suggests that automated evaluation, when properly guided, can serve as a credible and scalable proxy for expert review \cite{judgingLLMasJudge-Survey, 2024surveryLLMJudgesLI}. Although LLM-based judgments cannot fully replicate human reasoning, their consistency indicates potential for more auditable and self-monitoring AI systems in humanitarian analysis.

In conclusion, the results provide empirical evidence that large language models, when coupled with structured retrieval and evaluation mechanisms, can autonomously generate accurate, verifiable, and interpretable humanitarian situation reports. To support transparency and reproducibility, we render the code and prompts used in this work publicly available, while the dataset is available upon request (see Data Availability). By bridging the gap between information extraction and operational decision-making, the proposed framework advances both the scientific and practical understanding of how generative AI can support humanitarian action at scale. Incorporating continued expert feedback on content prioritization, structural flexibility, visualization features, and usability will be essential for developing next-generation systems capable of supporting deployment in complex crisis environments.

% All data supporting the findings of this study are openly available. This includes the datasets used as input to our framework, all prompt templates employed throughout the experiments, the annotated data produced during the evaluation, and the intermediate steps. All materials, together with the full source code required to reproduce the results, are accessible in the project repository at \url{https://github.com/Decostanzi/LLM-SituationalReport/}.

% A page is also online to visualize the generated reports: \url{https://decostanzi.github.io/LLM-SituationalReport/Viewer/viewer_v2.html}.

\section*{Data Availability}

The dataset used in this study was provided by Data Friendly Space (DFS) under a Memorandum of Understanding and includes copyrighted material from third-party humanitarian information platforms and news outlets. Due to licensing restrictions, the raw documents, annotations, and model outputs cannot be publicly released. Researchers interested in accessing the dataset may contact the corresponding authors, who will facilitate communication with DFS and support access requests in accordance with the data provider’s governance policy.

The source code, prompt templates, analysis scripts, and the final generated reports are openly available in the project repository at \url{https://github.com/idecost/LLM-SituationalReports}. Although the full texts cannot be shared due to licensing, links to the original, publicly viewable source documents are provided within the repository.
The generated reports can also be visualized via the online viewer at \url{https://idecost.github.io/LLM-SituationalReports/Viewer/viewer_v2.html}.

\section*{Acknowledgments}
The authors acknowledge support from the Lagrange Project of the ISI Foundation, funded by Fondazione
CRT.
K.K. thanks AECID (Spanish Agency for International Development Cooperation) for their support to data innovation
and Frontier Data Technologies through UNICEF’s Frontier Data Network.

\section*{Author contributions}

I.D., Y.M., and K.K. jointly conceived and designed the study.
I.D. developed the codebase, implemented the framework, and conducted the data analysis.
All authors contributed to interpreting the results, drafting sections of the manuscript, and revising it for important intellectual content.
Y.M. and K.K. supervised the project and provided methodological and domain expertise throughout the study.
All authors reviewed and approved the final manuscript.

\section*{Competing interests}
The author(s) declare no competing interests.

\section*{Funding}
This study was partially funded by Fondazione CRT through the Lagrange Project of the ISI Foundation. K.K. additionally received support from the Spanish Agency for International Development Cooperation (AECID) via UNICEF’s Frontier Data Network.

\clearpage

\end{document}